\documentclass[journal]{IEEEtran}
\ifCLASSINFOpdf
  \usepackage[pdftex]{graphicx}
\else
\fi

\usepackage{paralist}
\usepackage{booktabs}
\newcommand{\ra}[1]{\renewcommand{\arraystretch}{#1}}

\begin{document}
%
\title{Comparison of Time-Frequency Representations for Environmental Sound Classification using Convolutional Neural Networks}
%
%
%

\author{{Muhammad Huzaifah}
\thanks{M. Huzaifah (E0029863@u.nus.edu) is with the Arts and Creativity Lab, IDMI and NUS Graduate School for
Integrative Sciences and Engineering (NGS) at the National University of Singapore (NUS).}}

\maketitle

\begin{abstract}
Recent successful applications of convolutional neural networks (CNNs) to audio classification and speech recognition have motivated the search for better input representations for more efficient training. Visual displays of an audio signal, through various time-frequency representations such as spectrograms offer a rich representation of the temporal and spectral structure of the original signal. In this letter, we compare various popular signal processing methods to obtain this representation, such as short-time Fourier transform (STFT) with linear and Mel scales, constant-Q transform (CQT) and continuous Wavelet transform (CWT), and assess their impact on the classification performance of two environmental sound datasets using CNNs. This study supports the hypothesis that time-frequency representations are valuable in learning useful features for sound classification. Moreover, the actual transformation used is shown to impact the classification accuracy, with Mel-scaled STFT outperforming the other discussed methods slightly and baseline MFCC features to a large degree. Additionally, we observe that the optimal window size during transformation is dependent on the characteristics of the audio signal and architecturally, 2D convolution yielded better results in most cases compared to 1D.
\end{abstract}

\begin{IEEEkeywords}
signal processing, environmental sound classification, convolutional neural networks, deep learning.
\end{IEEEkeywords}

%
\IEEEpeerreviewmaketitle

\section{Introduction}
%
%
%
%
\IEEEPARstart{W}{hile} not receiving as much attention by the scientific community as speech processing tasks, environmental sound recognition nonetheless contributes to important applications in surveillance \cite {radhakrishnan2005audio}, robotics \cite{yamakawa2011environmental} and home automation \cite{wang2008robust} among others. In comparison to standard speech, environmental sounds are often more chaotic and noise-like, without the underlying phonetic structure that has been successfully modeled by traditional machine learning methods like the hidden Markov model (HMM). Recent work in this field has shown two distinctive developments: the utilization of deep neural networks (DNNs) in particular the convolutional neural network (CNN) as a classifier and feature extractor, and the use of the time-frequency representation of an audio signal, known as the spectrogram, as input.

CNN-based models were first adopted for speech recognition systems by Abdel-Hamid et al. \cite{abdel2014convolutional} for the TIMIT phone recognition task. This model was later improved architecturally in \cite{deng2013deep}, with added consideration to kernel size, pooling, network size and regularization, while other large-scale speech tasks were also carried out in \cite{sainath2015deep}\cite{sainath2013deep}\cite{lee2009unsupervised} using CNNs. More recently, Piczak \cite{piczak} and Salamon \cite{salamon} showed that a basic CNN could generally outperform existing methods for environmental sound classification provided sufficient data. 

To achieve desirable results, the classifier has to be paired with an appropriate input representation. Conventional choices were largely hand-crafted features such as Mel-frequency cepstral coefficients (MFCCs) or Perceptual Linear Prediction (PLP) coefficients that were previously state-of-the-art when used with Gaussian mixture model (GMM)-based HMMs. However, such cepstral features became less popular with deep learning algorithms as it was no longer essential for feature maps to be sufficiently de-correlated \cite{deng2013recent}\cite{deng2013new}. Conversely, the strength of CNNs lie with its ability to learn localized patterns through weight-sharing and pooling \cite{lecun1998gradient} - patterns present in the spectro-temporal features of spectrograms.   

In the domain of environmental sound it has been noted that time-frequency representations are especially useful as learning features \cite{orr2001speech}\cite{ghoraani2011time}\cite{khunarsal2013very}\cite{chu2009environmental}\cite{dennis2011spectrogram} due to the non-stationary and dynamic nature of the sounds. To extract these spectro-temporal features, a range of signal processing techniques have been proposed. A survey on environmental sound recognition by Chachada and Kuo \cite{chachada2013environmental} covers several methods, including sparse-representation-based techniques such as matching pursuit, power-spectrum-based techniques to obtain variants of the spectrogram, and several wavelet-based approaches. Another comparative study \cite{cowling2003comparison} investigated the performance of methods such as short-time Fourier transform (STFT), fast Wavelet transform (FWT) and continuous Wavelet transform (CWT) against stationary features like the aforementioned MFCC and PLP. The authors classified the extracted features using conventional machine learning techniques, including GMM-HMM, support vector machines (SVMs) and shallow artificial neural nets.

This letter builds upon previous comparative studies by focusing on the specifics of a CNN model as opposed to more traditional classifiers. We investigate four common approaches to obtain the time-frequency representation, namely the short-time Fourier transform (STFT) with both linear and Mel scales, the constant-Q transform (CQT) and the continuous Wavelet transform (CWT), while addressing additional considerations like window size. The impact of the different approaches is evaluated in comparison to baseline MFCC features on two publicly available environmental sound datasets (ESC-50, UrbanSound8K) through the classification performance of several CNN variants. 

\section{Experimental Methodology}

\subsection{Datasets}

The ESC-50 dataset \cite{esc50} comprises of 2000 short (5 second) environmental recording split equally among 50 classes. Classes were derived from 5 major groups: animals, natural soundscapes and water sounds, human non-speech sounds, interior/domestic sounds, and exterior/urban noises. The small number of samples coupled with a relatively large number of classes made this quite a demanding dataset for traditional classification methods. Previous work by Piczak \cite{piczak} showed that using a deep learning approach through CNNs markedly improved classification performance, yielding an accuracy of 64.5$\%$. A recent paper \cite{aytar2016soundnet} further improved this result to 74.2$\%$ using a deeper pre-trained network.

UrbanSound8K \cite{urbansound} is a collection of 8732 short (4 seconds or less) audio clips taken from field recordings. The dataset is divided into 10 distinct classes of  urban sounds: air conditioner, car horn, children playing, dog barking, drilling, engine idling, gun shot, jackhammer, siren and street music. Unlike ESC-50, the classes were not completely balanced, with the car horn, gun shot and siren sounds having fewer examples. The current sate-of-the-art on this dataset \cite{salamon} achieved a mean classification accuracy of 79.0$\%$.

\subsection{Pre-processing}

Proper pre-processing of the raw data was a major focus to make comparisons between the different transformations as fair as possible. Four main frequency-time representations were extracted in addition to MFCCs:
\begin{inparaenum}[\itshape a\upshape)]
\item linear-scaled STFT spectrogram
\item Mel-scaled STFT spectrogram
\item CQT spectrogram
\item CWT scalogram
\item MFCC cepstrogram.
\end{inparaenum} 

Firstly, all audio clips were standardized by padding/clipping to a 4 second duration on both datasets and resampled at 22050 Hz. Unlike \cite{piczak} and \cite{salamon}, whole clips were used for the subsequent transformations, including periods of silence and without additional augmentation. For STFT \cite{STFT}, the discrete Fourier transform with a sliding Hann window $w[n]$ was applied to overlapping segments of the signal, given by
\begin{equation}
X[n,k] = \sum\limits_{m=0}^{L-1} x[m].w[n-m].e^{-i \frac{2 \pi k}{N} m}
\label{STFT}
\end{equation}
Varying the window length $L$ results in a trade-off between frequency and time resolution. Both wideband ($L=2048$) and narrowband ($L=512$) transforms were used to probe this effect. Hop size was fixed at $L/2$  in both cases. 

\begin{figure}[h]
  \centering
     \includegraphics[width=0.22\textwidth]{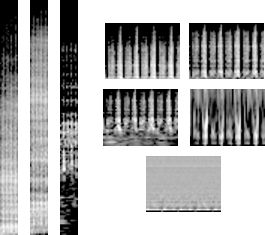}
  \caption{Examples of time-frequency spectrogram-like representations used as input extracted from the same ESC-50 sample (Handsaw 5-253094-c). Left, from far left: wideband linear-STFT, Mel-STFT, CQT. Right, clockwise from top right: narrowband Mel-STFT, CWT, MFCC, CQT, linear-STFT.}
  \label{spec}
\end{figure}

Spectrograms are defined as the squared magnitude of the STFT, giving the power of the sound for a particular frequency and time in the third dimension. The values were converted to a logarithmic scale (decibels) then normalized to [-1,1] generating a single-channel greyscale image (Fig.\ref{spec}). The frequency bins were either spaced linearly or mapped onto the Mel scale with 512 or 128 Mel bands for wideband and narrowband respectively.

The same general procedure was applied to the other transforms. The CQT \cite{CQT} is a bank of filters corresponding to tonal spacing where each filter is equivalent to a subdivision of an octave, with central frequencies given by $f_k = {(2^{\frac{1}{b}})}^k.f_{min}$. Here $f_k $ denotes the frequency of the $k^{th}$ spectral component and $b$ the number of filters per octave. As the name suggests, the $Q$ value, which is the ratio of central frequency to bandwidth, should be constant
\begin{equation} 
 Q = \frac{f_k}{\Delta f_k} = \frac{f_k}{f_{k+1}-f_k}= {(2^{\frac{1}{b}}-1)}^{-1}
\end{equation} 
Like the STFT, wideband ($k=1024,b=128$) and narrowband ($k=256,b=32$) versions of the CQT coefficients were extracted using
\begin{equation}
X[n,k] = \frac{1}{N[K]}. \sum\limits_{m=0}^{N[K]-1} x[n].W[k,n].e^{-i 2 \pi \frac{nQ}{N[k]}}
\label{CQT}
\end{equation}

Instead of decomposing a signal into sinusoids like the FT, the CWT uses a combination of basis functions that are located in both real and Fourier space i.e. frequency domain. Here the CWT was specified with 256 frequency bins and a Morlet mother function $\Psi_{(a,b)}$ that has been used in previous audio recognition studies \cite{orr2001speech}\cite{cowling2003comparison}
\begin{equation}
F(a,b) = <f,\Psi_{(a,b)}> = \int_{- \infty}^{+ \infty} \! f(x).\Psi_{(a,b)}^{*} \, \mathrm{d}x
\label{CWT}
\end{equation}
The CWT analogue of the spectrogram was attained by computing the squared value of the resultant Wavelet coefficients \cite{rioul1991wavelets}. Since the CWT allows for arbitrary time-frequency resolution limited only by sampling rate, only one transformation was carried out corresponding to narrowband dimensions.

Finally, MFCCs were obtained using the standard procedure and arranged as a cepstrogram. The coefficients were also normalized to [-1,1] but were not log-scaled. 

To keep the number of input feature maps (area of spectrogram) identical, the images were further downscaled to 37$\times$50 pixels for CWT, MFCC and all narrowband spectrograms, and 154$\times$12 pixels for wideband spectrograms, resulting in 1850 and 1848 input parameters respectively. Downscaling was done with PIL\footnotemark[1]\footnotetext[1]{http://pillow.readthedocs.io/en/4.0.x/handbook/overview.html} using Lanczos resampling for optimal results. The added effect of the resizing was a significant speed up in training times without sacrificing much accuracy. 

Audio processing was mostly carried out using librosa \cite{mcfee2017librosa} with the exception of CWT for which pyWavelets \cite{wasilewski2010pywavelets} was used. 

\subsection{Network Architecture and Evaluation}

\begin{figure}[h]
  \centering
     \includegraphics[width=0.40\textwidth]{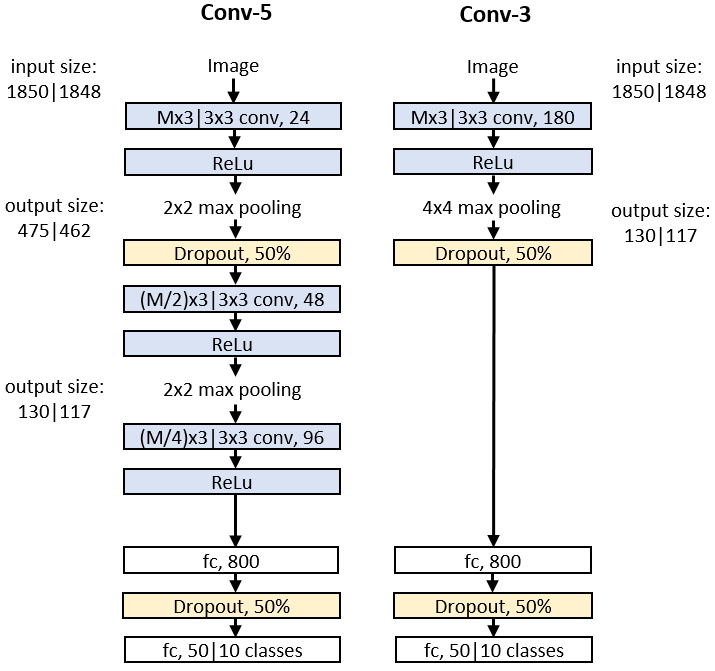}
  \caption{Network architecture of the deeper (Conv-5) and shallower (Conv-3) CNN models. 3$\times$3 and M$\times$3 sized filters were used in the convolutional layers. While containing more layers, Conv-5 had fewer number of parameters in each layer compared to Conv-3's single convolutional layer.}
  \label{cnn}
\end{figure}

Shallower and deeper variations of a CNN were implemented, informed by popular image recognition models and the CNNs in \cite{piczak}\cite{salamon}. The overall architecture is illustrated in Fig.\ref{cnn}.

Two types of convolutional filters were considered, a 3$\times$3 square filter and a M$\times$3 rectangular filter, spanning all M frequency bins, that essentially forces a one-dimensional convolution over time. As opposed to natural images where both axes contain spatial information, the two are nonsymmetric in a spectrogram. The invariant property may hold over time as a translation in a spectrogram image but it was not immediately evident whether pitch invariance will hold over the full frequency spectrum. For instance \cite{salamon} used a small filter, \cite{piczak} used one spanning just short of all the frequency bins while \cite{anand2015convoluted} implemented both types to varying success. Our results show that its performance is partly effected by the scale or transformation used.

The convolutional layers were interspersed with rectified linear unit (ReLu) and max pooling layers with stride sizes equal to the pooling dimensions. The Conv-3 employed more aggressive max pooling compared to the Conv-5. Dropout \cite{srivastava2014dropout} was utilized during training after the first convolutional and fully-connected layers, in addition to L2-regularization on all weight layers, to reduce overfitting .

Training was performed using Adam optimization \cite{kingma2014adam} with a batch size of 100, and cross-entropy for the loss function. Both datasets came prearranged into non-overlapping folds, and all models were evaluated using 5-fold (ESC-50) and 10-fold (UrbanSound8K) cross validation with a single fold held out as a test set for each round of validation while training with the remainder. Models were trained for 200 epochs for ESC-50 and 100 epochs for UrbanSound8K. The order of samples in the training and test sets were randomly shuffled after each training epoch. The reported results are median values of the test classification accuracy from the best training epoch across 4 separate cross validation runs for ESC-50 and 2 separate runs for UrbanSound8K. All network parameters were kept constant for each fold and run although the weights were randomly initialized each time following a truncated normal distribution. The network was implemented in Python with Tensorflow \cite{abadi2016tensorflow}.

\section{Results}

\begin{table*}[t]
\ra{1.1}
\caption{Median and median absolute deviation of accuracies ($\%$) for ESC-50 and UrbanSound8K}
\centerline{
\begin{tabular}{@{}rrrcrrcrrcrcr@{}}\toprule
& \multicolumn{2}{c}{Linear-STFT} & \phantom{a} & \multicolumn{2}{c}{Mel-STFT} &
\phantom{a} & \multicolumn{2}{c}{CQT} & \phantom{a} & \multicolumn{1}{c}{CWT} & \phantom{a} & \multicolumn{1}{c}{MFCC}\\
\cmidrule{2-3} \cmidrule{5-6} \cmidrule{8-9}
& $wideband$ & $narrowband$ &&$wideband$ & $narrowband$ && $wideband$ & $narrowband$\\ \midrule
\textbf{ESC-50}\\
Conv-5: M$\times$3 & 44.50$\pm$2.00&\textbf{46.62$\pm$2.25}&&\textbf{46.25$\pm$2.00}& \textbf{48.00$\pm$1.63}&&42.00$\pm$2.37&42.62$\pm$1.50&& 38.25$\pm$1.50&& 30.50$\pm$1.50\\
3$\times$3 & 49.25$\pm$0.75& 50.00$\pm$1.88&& \textbf{50.87$\pm$2.50}& \textbf{53.75$\pm$1.75}&& 46.87$\pm$1.13& 48.62$\pm$2.00&& 40.50$\pm$2.13&& 36.62$\pm$2.13\\
Conv-3: M$\times$3& 52.12$\pm$1.12&\textbf{55.12$\pm$1.88}&& \textbf{56.37$\pm$1.63}& \textbf{56.25$\pm$1.75}&& \textbf{54.37$\pm$2.25}& \textbf{53.50$\pm$1.87}&& 46.50$\pm$1.63&& 35.25$\pm$2.75\\
3$\times$3 & \textbf{55.00$\pm$1.37}& 53.00$\pm$1.62&& \textbf{54.00$\pm$1.25}& \textbf{55.00$\pm$1.63}&& 51.75$\pm$1.25& 51.62$\pm$2.25&& 46.62$\pm$1.87&& 35.00$\pm$0.75\\
\textbf{UrbanSound8K}\\
Conv-5: M$\times$3 & 61.19$\pm$4.81 & \textbf{63.44$\pm$3.39} && \textbf{62.22$\pm$5.19} & \textbf{64.97$\pm$3.69} && \textbf{62.87$\pm$3.25} & \textbf{63.12$\pm$3.25} && 56.90$\pm$2.10 && 59.23$\pm$3.24\\
3$\times$3 & \textbf{67.94$\pm$4.22} & 62.83$\pm$4.73 && \textbf{69.59$\pm$4.19} & \textbf{65.31$\pm$2.19} &&\textbf{69.25$\pm$4.69} & 64.33$\pm$3.60 && 61.56$\pm$1.80 && 57.15$\pm$1.81\\
Conv-3: M$\times$3 & \textbf{68.81$\pm$4.50} & 66.72$\pm$2.72&& \textbf{70.69$\pm$4.06} & \textbf{68.29$\pm$3.00}&&\textbf{70.94$\pm$4.06}& \textbf{67.06$\pm$3.12}&&64.00$\pm$2.17&& 64.87$\pm$2.17\\
3$\times$3 & \textbf{70.94$\pm$2.94}& 68.19$\pm$3.25&& \textbf{74.66$\pm$3.39}& \textbf{71.25$\pm$1.85}&& \textbf{73.03$\pm$3.56}& 68.31$\pm$2.35&& 64.75$\pm$1.44&& 62.81$\pm$4.03\\
\bottomrule
\end{tabular}
}
\end{table*}

\subsection{Impact of time-frequency representation}

Median classification accuracy and their corresponding median absolute deviations\footnotemark[2]\footnotetext[2]{Median was utilized as a statistical measure as opposed to mean for its robust property especially against outliers. Deviations were chiefly the result of the cross-validation protocol with each fold being a distinct test set, but also came from other stochastic factors such as the random initialization of weights and batching during optimization. For this reason, the whole k-fold cross validation process was repeated more than once.} for all experimental cases are presented in Table 1. It was immediately evident that classification performance with spectral representations as input outperformed traditional MFCC features. Other than the models with rectangular filters on the UrbanSound8K dataset for which MFCCs returned a better accuracy than CWT, this was consistent throughout. In fact time-frequency representations bettered MFCCs by a relatively wide margin, up to 15-20$\%$ in some cases. Even so, the confusion matrices in Fig.3 indicates that similar sets of classes were misclassified using MFCC and narrowband spectrogram features, albeit at a greater degree for the former. This suggests that the features provided by both transformations were closely related although spectral features were ultimately more discriminatory.    

\begin{figure}[h]
\centering
\begin{minipage}{.48\linewidth}
  \includegraphics[width=\linewidth]{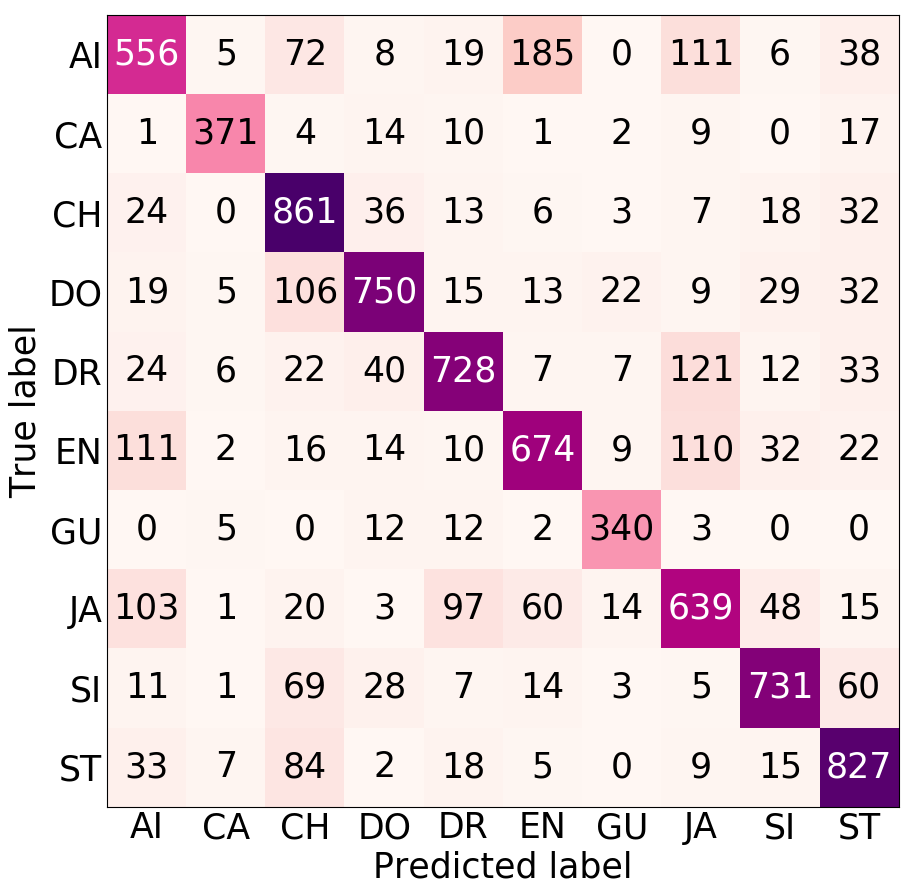}
  \label{img1}
\end{minipage}
\hspace{.01\linewidth}
\begin{minipage}{.48\linewidth}
  \includegraphics[width=\linewidth]{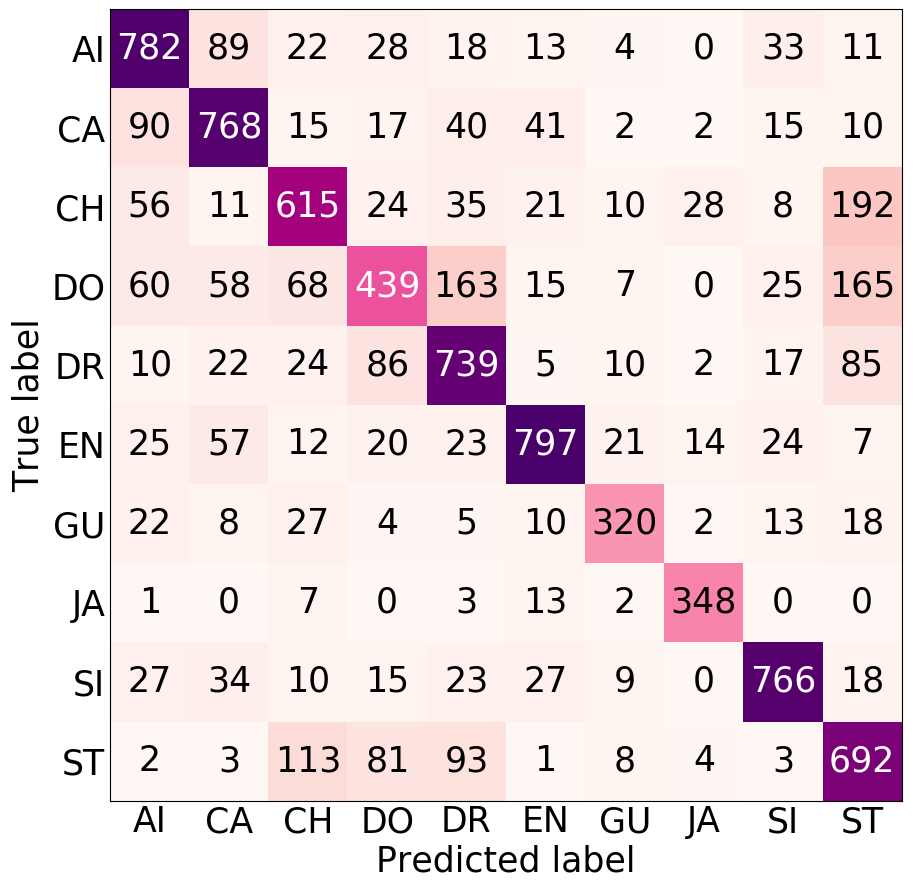}
  \label{img2}
\end{minipage}
\begin{minipage}{.48\linewidth}
  \includegraphics[width=\linewidth]{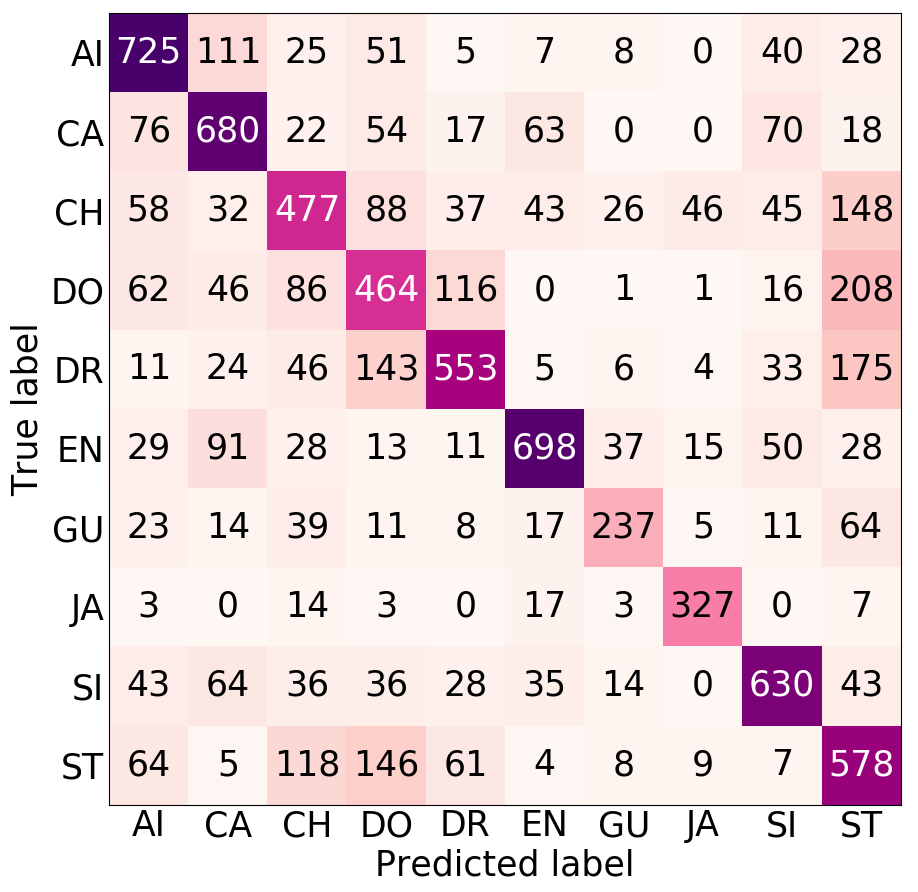}
  \label{img3}
\end{minipage}
\hspace{.01\linewidth}
\begin{minipage}[c]{.48\linewidth}
  \caption{Confusion matrices for the best performing Conv-3 model with 3$\times$3 filter on UrbanSound8K with different inputs: wideband (top left) and narrowband (top right) Mel-STFT spectrogram, MFCC cepstrogram (left). The wideband confusion matrix displays a similar distribution across classes as its analogue in \cite{piczak} and \cite{salamon}. }
\end{minipage}
\end{figure}

Among the spectral transformations under consideration, it was observed that linear-STFT, Mel-STFT and CQT performed comparably on both datasets. On the other hand, CWT results were lower and closer to MFCC, especially for UrbanSound8K. The top performers for each model variation (indicated in bold print) were determined by means of an ANOVA and post-hoc Tukey test\footnotemark[3]\footnotetext[3]{There were multiple top performers when the null hypothesis could not be rejected between pairs of transformations. Significance value used was 0.05.}. 

\subsection{Effect of CNN architecture and filter size}

Overall, the shallower Conv-3 model tended to yield better accuracies than Conv-5 regardless of input. A probable explanation is the diminishing returns of the deeper model due to significant overfitting. While it simplified the experimental methodology, using whole audio clips as input inevitably resulted in fewer training examples with less variation, impairing generalizability of the models. This can be seen when comparing the training and test curves (Fig.4) that tended to be closer together on Conv-3 than Conv-5.   

\begin{figure}[h]
\centering
\begin{minipage}{.48\linewidth}
  \includegraphics[width=\linewidth]{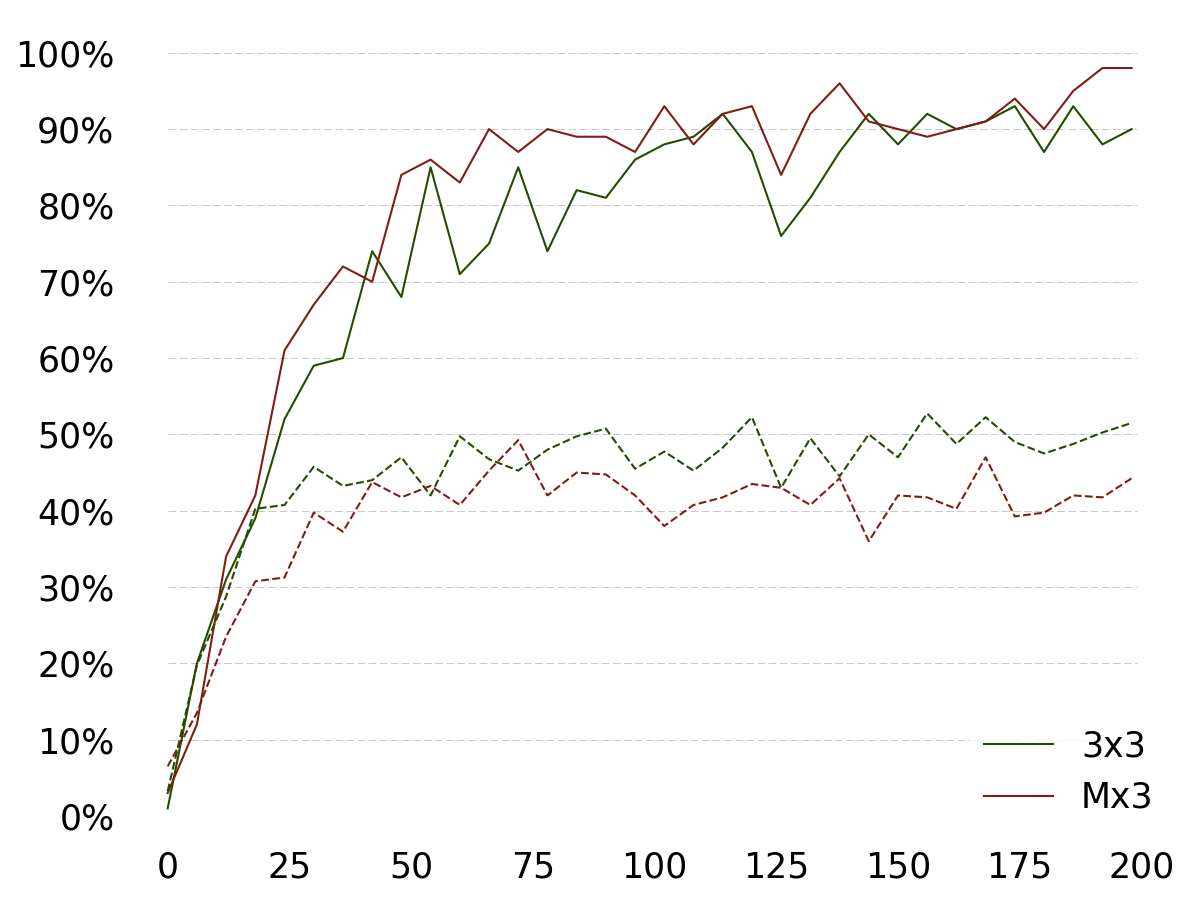}
  \label{img1}
\end{minipage}
\hspace{.01\linewidth}
\begin{minipage}{.48\linewidth}
  \includegraphics[width=\linewidth]{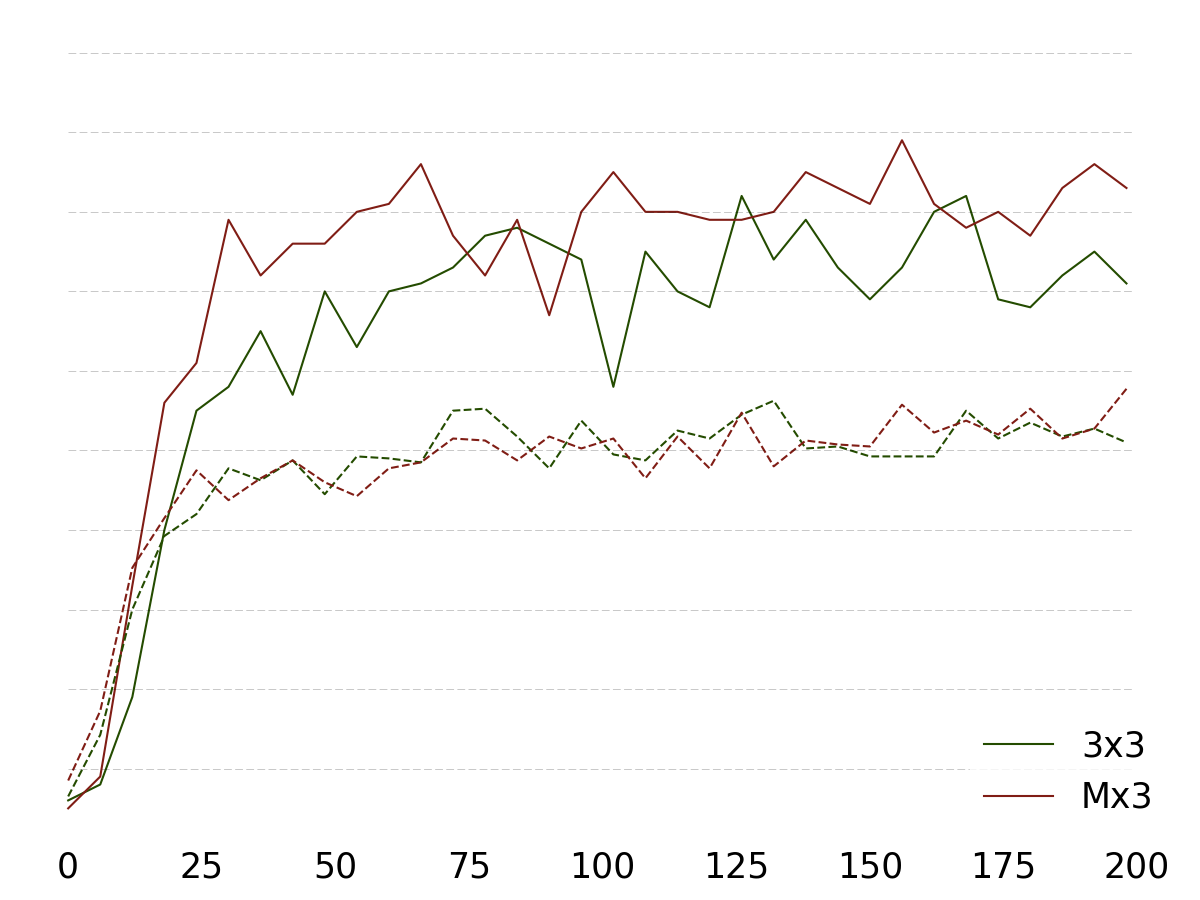}
  \label{img2}
\end{minipage}
  \caption{ESC-50 training (solid line) and test (dashed line) accuracy over epochs using Conv-5 (left) and Conv-3 (right) models with a narrowband Mel-STFT spectrogram as input.}
\end{figure}

Further, 2D convolution generally gave better results than 1D with the notable anomaly of Conv-3 for ESC-50. Again, this could be partly attributed to overfitting of the bigger M$\times$3 filters as illustrated in Fig.4, although it cannot be discounted that some invariant properties of the frequencies were captured by the smaller filter. Theoretically, Mel-STFT and especially CQT was expected to be more tolerant to pitch invariance, the former being based on a log-based perceptual scale of pitches and the latter preserving harmonic structure despite changes to the fundamental frequency. This may have been the case seeing the accuracy gains moving from M$\times$3 to 3$\times$3 filters, although other transforms also displayed similar improvements. We posit that CQT may be even more beneficial for music analysis where timbre signatures are more evident in the instrumentation as compared to environmental sounds.  

\subsection{Wideband vs Narrowband}
 
The benefit of wideband against narrowband transforms were not consistent across both datasets. This may be indicative of a disparity in the types of environmental sounds present in both datasets. More interestingly perhaps, comparing the confusion matrices reveals that each specializes in discriminating certain classes of sound. The wideband Mel-STFT fared poorly for short temporal sounds like ``drilling" or ``jackhammer" and for droning sounds like ``air conditioner" but excelled at classes with high frequency variations such as ``children playing". For narrowband, having better temporal resolution but poorer frequency resolution, the inverse was true.  

\section{Conclusion}

The main objective of this paper was to perform a comparative study between different forms of commonly used time-frequency representations and evaluate their impact on the CNN classification performance of environmental sound data. It was shown that Mel-STFT spectrograms were consistently good performers across the variations tested, although Linear-STFT and CQT also did well on some models. Generally all time-frequency representations produced better accuracies than baseline MFCC features, corroborating previous studies. The effectiveness of using a wide or narrow window during the transformation, if applicable, was determined to be class dependent. Further insight into the audio characteristics of the datasets would help in determining which variation would be more advantageous. Finally, we considered capability of both 2D and 1D convolution over time by changing the filter size. 2D convolution seemed to work better for most cases, with the exception of the shallower model on the ESC-50 dataset. Nonetheless, the best approach may instead lie in between, such as by using a variable-sized filter, to properly trade-off between invariance to pitch and better discriminative power.   

\section*{Acknowledgment}

The author would like to thank Haifa Beji and Lonce Wyse for their input and feedback.



\ifCLASSOPTIONcaptionsoff
  \newpage
\fi



\bibliographystyle{IEEEtran}
\bibliography{ieeebib}

\begin{thebibliography}{10}
\providecommand{\url}[1]{#1}
\csname url@samestyle\endcsname
\providecommand{\newblock}{\relax}
\providecommand{\bibinfo}[2]{#2}
\providecommand{\BIBentrySTDinterwordspacing}{\spaceskip=0pt\relax}
\providecommand{\BIBentryALTinterwordstretchfactor}{4}
\providecommand{\BIBentryALTinterwordspacing}{\spaceskip=\fontdimen2\font plus
\BIBentryALTinterwordstretchfactor\fontdimen3\font minus
  \fontdimen4\font\relax}
\providecommand{\BIBforeignlanguage}[2]{{%
\expandafter\ifx\csname l@#1\endcsname\relax
\typeout{** WARNING: IEEEtran.bst: No hyphenation pattern has been}%
\typeout{** loaded for the language `#1'. Using the pattern for}%
\typeout{** the default language instead.}%
\else
\language=\csname l@#1\endcsname
\fi
#2}}
\providecommand{\BIBdecl}{\relax}
\BIBdecl

\bibitem{radhakrishnan2005audio}
R.~Radhakrishnan, A.~Divakaran, and A.~Smaragdis, ``Audio analysis for
  surveillance applications,'' in \emph{Applications of Signal Processing to
  Audio and Acoustics, 2005. IEEE Workshop on}.\hskip 1em plus 0.5em minus
  0.4em\relax IEEE, 2005, pp. 158--161.

\bibitem{yamakawa2011environmental}
N.~Yamakawa, T.~Takahashi, T.~Kitahara, T.~Ogata, and H.~Okuno, ``Environmental
  sound recognition for robot audition using matching-pursuit,'' \emph{Modern
  Approaches in Applied Intelligence}, pp. 1--10, 2011.

\bibitem{wang2008robust}
J.-C. Wang, H.-P. Lee, J.-F. Wang, and C.-B. Lin, ``Robust environmental sound
  recognition for home automation,'' \emph{IEEE transactions on automation
  science and engineering}, vol.~5, no.~1, pp. 25--31, 2008.

\bibitem{abdel2014convolutional}
O.~Abdel-Hamid, A.-r. Mohamed, H.~Jiang, L.~Deng, G.~Penn, and D.~Yu,
  ``Convolutional neural networks for speech recognition,'' \emph{IEEE/ACM
  Transactions on audio, speech, and language processing}, vol.~22, no.~10, pp.
  1533--1545, 2014.

\bibitem{deng2013deep}
L.~Deng, O.~Abdel-Hamid, and D.~Yu, ``A deep convolutional neural network using
  heterogeneous pooling for trading acoustic invariance with phonetic
  confusion,'' in \emph{Acoustics, Speech and Signal Processing (ICASSP), 2013
  IEEE International Conference on}.\hskip 1em plus 0.5em minus 0.4em\relax
  IEEE, 2013, pp. 6669--6673.

\bibitem{sainath2015deep}
T.~N. Sainath, B.~Kingsbury, G.~Saon, H.~Soltau, A.-r. Mohamed, G.~Dahl, and
  B.~Ramabhadran, ``Deep convolutional neural networks for large-scale speech
  tasks,'' \emph{Neural Networks}, vol.~64, pp. 39--48, 2015.

\bibitem{sainath2013deep}
T.~N. Sainath, A.-r. Mohamed, B.~Kingsbury, and B.~Ramabhadran, ``Deep
  convolutional neural networks for lvcsr,'' in \emph{Acoustics, speech and
  signal processing (ICASSP), 2013 IEEE international conference on}.\hskip 1em
  plus 0.5em minus 0.4em\relax IEEE, 2013, pp. 8614--8618.

\bibitem{lee2009unsupervised}
H.~Lee, P.~Pham, Y.~Largman, and A.~Y. Ng, ``Unsupervised feature learning for
  audio classification using convolutional deep belief networks,'' in
  \emph{Advances in neural information processing systems}, 2009, pp.
  1096--1104.

\bibitem{piczak}
K.~J. Piczak, ``Environmental sound classification with convolutional neural
  networks,'' in \emph{Machine Learning for Signal Processing (MLSP), 2015 IEEE
  25th International Workshop on}.\hskip 1em plus 0.5em minus 0.4em\relax IEEE,
  2015, pp. 1--6.

\bibitem{salamon}
J.~Salamon and J.~P. Bello, ``Deep convolutional neural networks and data
  augmentation for environmental sound classification,'' \emph{IEEE Signal
  Processing Letters}, vol.~24, no.~3, pp. 279--283, 2017.

\bibitem{deng2013recent}
L.~Deng, J.~Li, J.-T. Huang, K.~Yao, D.~Yu, F.~Seide, M.~Seltzer, G.~Zweig,
  X.~He, J.~Williams \emph{et~al.}, ``Recent advances in deep learning for
  speech research at microsoft,'' in \emph{Acoustics, Speech and Signal
  Processing (ICASSP), 2013 IEEE International Conference on}.\hskip 1em plus
  0.5em minus 0.4em\relax IEEE, 2013, pp. 8604--8608.

\bibitem{deng2013new}
L.~Deng, G.~Hinton, and B.~Kingsbury, ``New types of deep neural network
  learning for speech recognition and related applications: An overview,'' in
  \emph{Acoustics, Speech and Signal Processing (ICASSP), 2013 IEEE
  International Conference on}.\hskip 1em plus 0.5em minus 0.4em\relax IEEE,
  2013, pp. 8599--8603.

\bibitem{lecun1998gradient}
Y.~LeCun, L.~Bottou, Y.~Bengio, and P.~Haffner, ``Gradient-based learning
  applied to document recognition,'' \emph{Proceedings of the IEEE}, vol.~86,
  no.~11, pp. 2278--2324, 1998.

\bibitem{orr2001speech}
M.~C. Orr, D.~S. Pham, B.~Lithgow, and R.~Mahony, ``Speech perception based
  algorithm for the separation of overlapping speech signal,'' in
  \emph{Intelligent Information Systems Conference, The Seventh Australian and
  New Zealand 2001}.\hskip 1em plus 0.5em minus 0.4em\relax IEEE, 2001, pp.
  341--344.

\bibitem{ghoraani2011time}
B.~Ghoraani and S.~Krishnan, ``Time--frequency matrix feature extraction and
  classification of environmental audio signals,'' \emph{IEEE transactions on
  audio, speech, and language processing}, vol.~19, no.~7, pp. 2197--2209,
  2011.

\bibitem{khunarsal2013very}
P.~Khunarsal, C.~Lursinsap, and T.~Raicharoen, ``Very short time environmental
  sound classification based on spectrogram pattern matching,''
  \emph{Information Sciences}, vol. 243, pp. 57--74, 2013.

\bibitem{chu2009environmental}
S.~Chu, S.~Narayanan, and C.-C.~J. Kuo, ``Environmental sound recognition with
  time--frequency audio features,'' \emph{IEEE Transactions on Audio, Speech,
  and Language Processing}, vol.~17, no.~6, pp. 1142--1158, 2009.

\bibitem{dennis2011spectrogram}
J.~Dennis, H.~D. Tran, and H.~Li, ``Spectrogram image feature for sound event
  classification in mismatched conditions,'' \emph{IEEE Signal Processing
  Letters}, vol.~18, no.~2, pp. 130--133, 2011.

\bibitem{chachada2013environmental}
S.~Chachada and C.-C.~J. Kuo, ``Environmental sound recognition: A survey,'' in
  \emph{Signal and Information Processing Association Annual Summit and
  Conference (APSIPA), 2013 Asia-Pacific}.\hskip 1em plus 0.5em minus
  0.4em\relax IEEE, 2013, pp. 1--9.

\bibitem{cowling2003comparison}
M.~Cowling and R.~Sitte, ``Comparison of techniques for environmental sound
  recognition,'' \emph{Pattern recognition letters}, vol.~24, no.~15, pp.
  2895--2907, 2003.

\bibitem{esc50}
K.~J. Piczak, ``Esc: Dataset for environmental sound classification,'' in
  \emph{Proceedings of the 23rd ACM international conference on
  Multimedia}.\hskip 1em plus 0.5em minus 0.4em\relax ACM, 2015, pp.
  1015--1018.

\bibitem{aytar2016soundnet}
Y.~Aytar, C.~Vondrick, and A.~Torralba, ``Soundnet: Learning sound
  representations from unlabeled video,'' in \emph{Advances in Neural
  Information Processing Systems}, 2016, pp. 892--900.

\bibitem{urbansound}
J.~Salamon, C.~Jacoby, and J.~P. Bello, ``A dataset and taxonomy for urban
  sound research,'' in \emph{Proceedings of the 22nd ACM international
  conference on Multimedia}.\hskip 1em plus 0.5em minus 0.4em\relax ACM, 2014,
  pp. 1041--1044.

\bibitem{STFT}
J.~B. Allen and L.~R. Rabiner, ``A unified approach to short-time-fourier
  analysis and synthesis,'' \emph{Proceedings of the IEEE}, vol.~65, no.~11,
  November 1977.

\bibitem{CQT}
J.~C. Brown, ``Calculation of a constant q spectral transform,'' \emph{The
  Journal of the Acoustical Society of America}, 1991.

\bibitem{rioul1991wavelets}
O.~Rioul and M.~Vetterli, ``Wavelets and signal processing,'' \emph{IEEE signal
  processing magazine}, vol.~8, no.~4, pp. 14--38, 1991.

\bibitem{mcfee2017librosa}
\BIBentryALTinterwordspacing
B.~McFee, M.~McVicar, C.~Raffel, D.~Liang, O.~Nieto, J.~Moore, D.~Ellis,
  D.~Repetto, P.~Viktorin, and J.~F. Santos, ``Librosa: v0.5.0,'' 2017.
  [Online]. Available: \url{http://doi.org/10.5281/zenodo.293021}
\BIBentrySTDinterwordspacing

\bibitem{wasilewski2010pywavelets}
\BIBentryALTinterwordspacing
F.~Wasilewski, ``Pywavelets: Discrete wavelet transform in python,'' 2010.
  [Online]. Available: \url{http://www.pybytes.com/pywavelets}
\BIBentrySTDinterwordspacing

\bibitem{anand2015convoluted}
N.~Anand and P.~Verma, ``Convoluted feelings convolutional and recurrent nets
  for detecting emotion from audio data,'' in \emph{Technical Report}.\hskip
  1em plus 0.5em minus 0.4em\relax Stanford University, 2015.

\bibitem{srivastava2014dropout}
N.~Srivastava, G.~E. Hinton, A.~Krizhevsky, I.~Sutskever, and R.~Salakhutdinov,
  ``Dropout: a simple way to prevent neural networks from overfitting.''
  \emph{Journal of Machine Learning Research}, vol.~15, no.~1, pp. 1929--1958,
  2014.

\bibitem{kingma2014adam}
D.~Kingma and J.~Ba, ``Adam: A method for stochastic optimization,''
  \emph{arXiv preprint arXiv:1412.6980}, 2014.

\bibitem{abadi2016tensorflow}
M.~Abadi, A.~Agarwal, P.~Barham, E.~Brevdo, Z.~Chen, C.~Citro, G.~S. Corrado,
  A.~Davis, J.~Dean, M.~Devin \emph{et~al.}, ``Tensorflow: Large-scale machine
  learning on heterogeneous distributed systems,'' \emph{arXiv preprint
  arXiv:1603.04467}, 2016.

\end{thebibliography}
\end{document}